\documentclass[12pt]{article}
\usepackage{geometry}                % See geometry.pdf to learn the layout options. There are lots.
\usepackage[all]{xy}
\usepackage{graphicx}
\usepackage{amssymb}
\usepackage{epstopdf}
\begin{document}
 \newcommand{\bfz}{\textbf{0}}
  \newcommand{\bfu}{\textbf{1}}
  \newcommand{\mR}{\mathcal{R}}
   \newcommand{\mI}{\mathcal{I}}
    \newcommand{\I}{\mathcal{I}}
 \newtheorem{definition}{Definition}
\newtheorem{lemma}{Lemma}
\newtheorem{theorem}{Theorem}
\newtheorem{remark}{Remark}
\newtheorem{example}{Example}

\newpage
\thispagestyle{empty}
\author{Vittorio Cafagna and Gianluca Caterina}
  \title{Notes on a model for fuzzy computing}
    \maketitle

\abstract{In these notes we propose a setting for {\sl fuzzy computing} in a framework similar to that of well-established theories of computation: boolean, and quantum computing. Our efforts have been directed towards stressing the formal similarities: there is a common pattern underlying these three theories. We tried to conform our approach, as much as possible, to this pattern. This work was part of a project jointly with Professor Vittorio Cafagna. Professor Cafagna passed away unexpectedly in 2007. His intellectual breadth and inspiring passion for mathematics is still very well alive.}\footnote{The support of Professor Antonio Di Nola is gratefully acknowledged.}

\section{Introduction}
The most widely studied theoretical models of computation, namely classical computation (CC) and, in more recent years, quantum computing (QC), have in common a nice description in terms of their linear nature. 
There exists a `logic' associated to each of the above models, which is, very roughly speaking, the theory of the properties of the language of that model. As an example, boolean logic is associated to the boolean computation, which is still nowadays the theoretic model on which actual computing machines work. 

The idea behind a `fuzzy' logic is a very natural one, having been first conceived as an extension of the values of truth of a formula from a discrete set to a continuous one, and motivated by L. Zadeh's seminal paper \cite{Zah} ; however, for reasons that are related in part  to the non linear nature of fuzzy logic, the computational aspect of such a model has always  been quite elusive of a systematic investigation. \\
This work  is a first attempt to propose a `linear'  theory of fuzzy computing (FC) in analogy with CQ and QC, trying at the same to highlight the original algebraic characteristics of the fuzzy model with respect to the other models.

The motivation leading us toward this approach is that the language of matrices has turned to be very useful in theory of computation: in a linear model, a computation is just a linear map (i.e., a matrix) between two sets where the information is encoded, and most of the effort in computing theory is devoted to understand how to optimize the factorization of such matrices. In CQ elementary information is coded in a two-state entity, called simply {\sl bit}, while in QC the bit is replaced by the {\sl qubit}, which, as we shall discuss, encode information is a completely different way. In Fuzzy Computing, we have been searching for a structure in which to encode elementary fuzzy information, and we propose a model which has seemed to us the natural extension of the classical and quantum point of view.
 
In what follows, we will define our synoptic frame by reviewing the basic linear algebra on which CQ and QC rely on; we refer to the excellent treatises by R. Feynman \cite{fe1}, \cite{fe2} for the basic notions of  theory of computation (concepts like Turing machines, logic gates, etc.) and an exposition of the ideas that Quantum Computation has been built on.

We start our study by showing how linearity arises in classical and quantum computing; to use a more suggestive analogy, we want to see how a certain geometry (that is, an algebraic subset of some vector space along with a group-or some weaker structure- of linear transformations for which the given subset is invariant) can be associated to a model of computation.

\section{Classical Computation}

Turing machines (see Turing's seminal paper \cite{tu}) are the fundamental theoretical models of computation. However, real computer computers are finite in size, while for Turing machines we need a computer of infinite size.

The model we are going to present is the $circuit\ model$, equivalent to the Turing machines in terms of computational power, but more convenient for practical applications. In particular, it allows an unifying approach to Quantum, and as we shall see, to Fuzzy Computing.

Elementary information is represented by an object called $bit$, an independent variable which can take  only the values (also called `states`) 0 or 1.
 
\subsection{Classical gates}
A $circuit$ is composed by two fundamental components: $gates$ and $wires$, on which bits of information are stored and processed.\\
 \begin{definition}
 A $gate$ is a function $\{0,1\}^k\longrightarrow \{0,1\}^l$ from some 
number $k$ of $input\  bits$ to $l$ $output\ bits.$
\end{definition}
\begin{example}
The gate $NOT$ computes the function $f:\{0,1\}\longrightarrow\{0,1\}$, $x\mapsto x\oplus 1$, where $\oplus$ is the sum modulo 2.
\end{example}
There are many other useful elementary gates; $OR$, $AND$, $XOR$, $NAND$, $NOR,\\$ $FANOUT$ are some of them (let us remind that the latter implements the function $f:\{0,1\}\longrightarrow\{0,1\}^2$, $x\mapsto (x,x)$).\\

The following result is a milestone on theory of computation:
\begin{theorem}
Any function $f:\{0,1\}^n\longrightarrow\{0,1\}^m$ can be implemented using only $AND, XOR,
 and\ NOT$ gates.
\end{theorem}

$Proof$.\ We give the proof tor the case $f:\{0,1\}^n\rightarrow\{0,1\}$, which generalizes easily. The proof goes by induction on $n$. For $n=1$ there are $4$ possible cases: the identity, which is a circuit with a single wire; the flip, which we have seen it can be implemented by the NOT gate; the function which overwrites 0, which can be obtained by ANDing the input with an ancilla bit initialized at the state 0; and the function which overwrites 1, which can be implemented by ORing the input with an ancilla bit initialized at the state 1. Therefore the base of the induction has been  proved; in order to complete the proof, suppose that any function on $n$ bits can be computed by a circuit using only AND, XOR and NOT, and let $f$ a Boolean function on $n+1$ input bits. We can define $f_0$ and $f_1$ to be two $n-$bits functions by 
\[f_0(x_1,\dots,x_n)=f(0,x_1,\dots,x_n) \ \ \ \ f_1(x_1,x_2,\dots,x_n)=f(1,x_1,\dots,x_n)\]
These are, by induction computed using only the few gates mentioned above; a simple argument show that, once $f_0$ and $f_1$ have been computed, a circuit consisting of two AND, one NOT and a XOR can be implemented to finally compute $f$.
\begin{flushright}
$\square$
\end{flushright}

This result tells us that there exists a very small set of operations which are $universal$, that is, they can be used to build an arbitrary large computation.

The gates listed above have a $non\ invertible$ nature: given the output bits we cannot recover the input bits, which means that some information has been destroyed during the computation. 
In their work \cite{ft} Toffoli and Fredkin  have shown that any non invertible computation can be simulated by a reversible one (up to increasing the number of bits -$ancilla\ bits$- used in the computation), so that we have another version of Thm.1:

\begin{theorem}
Any function $f:\{0,1\}^n\longrightarrow\{0,1\}^m$ can be simulated by an invertible function $\tilde{f}:\{0,1\}^N\longrightarrow\{0,1\}^N$ for $N$ large enough.
\end{theorem}

Any invertible function  from a set $S$ of size $n$ in itself is represented by an element of the symmetric group on $n$ elements, therefore, by Thm.2, a computation can be regarded in a pure algebraic way, which we illustrate with the following basic example.\\

\begin{example}
If we think of $0$ and $1$ as elements of $\mathbb{R}$, we see that the map $\phi:\mathbb{R}\longrightarrow\mathbb{R}$, $x\mapsto 1-x$ 
 restricts to an involution on $[0,1]$ which swaps 0 with 1.
Let us now represent the states 0 and 1 as elements of the plane $\mathbb{R}^2$: we associate to 0 and 1 the vectors $|0\rangle$ and $|1\rangle$ 
  \[|0\rangle=(1,0)\ \  \textrm{and}\ \  |1\rangle =(0,1)\]
Consider the following map:
 \[\tilde{\phi}:\mathbb{R}^2\longrightarrow\mathbb{R}^2\]
 \[\tilde{\phi}: |x\rangle\mapsto \left(\begin{array}{cc}0 & 1 \\1 & 0\end{array}\right)|x\rangle\]

The effect of $\tilde{\phi}$ is clearly to swap $|0\rangle$ with $|1\rangle$.

\end{example}
The difference between $\phi$ and $\tilde{\phi}$ is that $\tilde{\phi}$ is linear whereas $\phi$ is not: adding a dimension allows us to linearize the $NOT$ operation.\\

$\{0,1\}$ is the most of fundamental example of a $Boolean\ algebra$.

\begin{definition}
A Boolean algebra is a lattice $(A,\wedge,\vee)$ (considered as an algebraic structure) with the following four additional properties:
\begin{enumerate}
\item bounded below: There exists an element 0, such that $a\vee  0 = a$ for all $a\in A$.
\item bounded above: There exists an element 1, such that $a\wedge1 = a$ for all $a \in A$.
\item distributive law: For all $a, b, c \in A, (a\vee b)\wedge  c = (a\wedge  c)  (b\vee  c)$.
\item existence of complements: For every a in A there exists an element $Âa \in A$ such that $a\vee  Âa = 1$ and $a\wedgeÂa=0.$

\end{enumerate}
\end{definition}
The map $\tilde{\phi}$ just defined is the only non trivial, invertible linear map $\mathbb{R}^2\rightarrow \mathbb{R}^2$ (different from the identity) from $\mathcal{B}(2)=\{|0\rangle,|1\rangle\}$ to itself.
$\tilde{\phi}$, which is the non trivial element of $S_2$, the permutation group on two elements,  represents the invertible version of the $NOT$ gate.\\

By Thm. 2, we can restrict ourselves to invertible maps from $\mathcal{B}(2)$ in itself, and conclude that, for a single bit,  classical computation corresponds to the geometry
\[( \mathcal{B}(2), S_2)\]
where $S_2$ is the group of permutation on $2$ objects.

Notice that the first component of the pair represents the subset of $\mathbb{R}^2$ where the information is stored, namely $\mathcal{B}(2)=\{|0\rangle,|1\rangle\}$  and the second the group of linear invertible transformations for which that set is invariant.
\subsection{Multiple bits} 
In order to perform computations that need more then just one single bit, we need to define an object that naturally encodes the concept of `ensemble' of bits :

\begin{definition}
Given $n$ bits $b_0,b_1,\dots,b_{n-1}$ we define its state space to be 
\[ \mathcal{B}(2^n)=\{|00\dots 0\rangle,|00\dots1\rangle,\dots,|11\dots1\rangle\}=\]
where $|b_{n-1},b_n,\dots,b_0\rangle$ represents the $(\sum_{i=0}^{n-1}2^ib_i)^{th}$ vector of the canonical basis of $\mathbb{R}^{2^n}$.
\end{definition}

Suppose now we have a system $S$ in a state represented by $|b_0b_1\dots b_k\rangle$ and another system $S'$ represented by $|b'_0b'_1\dots b'_{k'}\rangle$; then we can compute the state of the coupled system $S,S'$ by taking the tensor product of the states:

\[|b_0b_1\dots b_k\rangle\otimes|b'_0b'_1\dots b'_{k'}\rangle=|b_0b_1\dots b_kb'_0b'_1\dots b'_{k'}\rangle\]

We remark the fact that, since we have represents all the possible states of a finite number of bits as vectors over the real numbers, the tensor product we are considering is the ordinary tensor product of vectors over the reals: to any configuration (ensemble of bits). The group of invertible maps on $\mathcal{B}(2^n)$ is the symmetric group $S_{2^n}$ that can be represented by the group of permutation matrices $P(2^n)$; therefore we can conclude that 
\begin{theorem}
The geometry of classical computation is given, by 
\[(\mathcal{B}(2^n),P(2^n))\]
\end{theorem}
\begin{flushright}
$\square$
\end{flushright}

\section{Non deterministic computation}

Suppose a bit is stored on a ordinary hard drive. The initial state of the bit is 0 or 1, but after a while the changes in the magnetic fields can cause the bit to scramble, in some case until the flipping of the initial state. We can model this situation by assigning a probability $p$ for the bit to flip, and its complement $1-p$ for the bit to stay the same.

The reason why we can do that successfully is because there is a model, namely Maxwell's equation that can predict precisely enough how the magnetic fields interact with the hard disk, from which we can deduce a the probability $p$ that a flip in the state of the bit will occur in a given interval of time.

In summary, a bit will be in the state $|0\rangle$ with probability $p$ and in the state $|1\rangle$ with probability $1-p$; this leads to the following definition:
\begin{definition}
\begin{itemize}
\item A stochastic elementary bit is an element of
\[Pr(2)=\{(p_1,p_2)\in\mathcal{I}^2|p_1+p_2=1\}\] 
\item $St(2)$ the set of linear invertible matrices with entries in $\mathcal{I}$ preserving $Pr(2)$
\end{itemize}
\end{definition}
where  $\mathcal{I}=[0,1]$.

\begin{lemma}
If $M\in St(2)$ then $M$ is of the form
\[M=\{\left(\begin{array}{cc}x & y \\1-x & 1-y\end{array}\right)|(x,y)\in\mathcal{I}^2\}\]
\end{lemma}
$Proof.$
Suppose $\alpha+\beta=1$, $\alpha,\beta\in\mathcal{I}$ and $x,y,w,z\ \in\mathcal{I}$ Then we have:
\[\left(\begin{array}{cc}x & y \\ w & z\end{array}\right)\left(\begin{array}{c} \alpha \\\beta \end{array}\right)=\left(\begin{array}{c} x\alpha +y\beta  \\ w\alpha+z\beta\end{array}\right)\]
Then we notice that
 \[x\alpha +y\beta+w\alpha+z\beta=\alpha(x+w)+\beta(y+z)=1\Rightarrow (x+w)=(y+z)=1\]
 \begin{flushright}
$\square$\
\end{flushright}

The higher dimensional generalization is straightforward:
\begin{definition}
Given $n$ stochastic bits we define its state space to be $Pr(2^n)$ where 
\[Pr(2^n)=\{(x_0,x_1,\dots x_{2^n-1})\in\mathcal{I}^{2^n}|\sum_{i=0}^{2^n-1}x_i=1\}\]
and the set of $n\times n$ stochastic matrices to be $St(2^n)$, where
\[St(2^n)=\{\left(\begin{array}{cccc}a_{11} & a_{12} & \dots & a_{12^n} \\a_{21} & a_{22} & \dots & a_{22^n} \\\dots &  &  & \dots \\a_{2^n1} &  &  & a_{2^n2^n}\end{array}\right)|\sum_{i=1}^{2^n}a_{ki}=1\ \forall k\in\{1,2,\dots 2^n\}\}\]
\end{definition}

Interestingly, $St(2^n)$ is a semigroup, not a group. Indeed, although $St(2^n)$ is closed under ordinary matrix multiplication, the inverse of a given stochastic matrix does not have to be necessary stochastic.

The main application of the machinery introduced in this section is the control of the noise in classical systems. Suppose indeed that we have been given faulty components to build the circuit. Our circuit consiste of a single input bit $x$, to which are applied, say, two consecutive (faulty) $NOT$ gates, producing an intermediate bit $y$ and a final outpot bit $z$. It seems reasonable to assume that the probability that the second $NOT$ works correctly is independent on the probability that the first $NOT$ works correctly. This assumption makes such a  `computation' a special kind of {\sl stochastic process}, known as {\sl Markov process}.

\section{Quantum computing}
One of the most recent and growing field in information theory goes under the name of Quantum Computing, whose birth was marked by  the seminal paper by Richard Feynman \cite{fe1}, in which the possibility of a parallel `quantum' computing' is conjectured. The main idea is that quantum interference phenomena, and the properties of the basic quantum entities (especially the possibility of living in a `superposition' of classical states), can improve dramatically the efficiency of the computation, if an hardware with sub-atomic hardwares will ever be build (just recently there are claims that a 8-bits quantum computer has been build).

In order to translate this model in an algebraic framework, we start by introducing the $qubit$, which is the elementary entity where the information can be encoded; let us first introduce the {\sl state space} for `quantum' bits:
\begin{definition} 
A {\sl qubit} is a point on the following algebraic curve in $\mathbb{C}^2$:
\[S_1(\mathbb{C})=\{(x_1,x_2)\in\mathbb{C}^2|\ \ |x_1|^2+|x_2|^2=1\}\]
\end{definition}

\begin{lemma}
The group of invertible $2\times 2$ matrices acting on  $S_1(\mathbb{C})$ is the Unitary group of $2\times 2$ matrices $U(2)$.
\end{lemma}
$Proof.$ The proof is analogue to that of lemma 5.
\begin{flushright}
$\square$\
\end{flushright}

\begin{definition}
Given $a,b,c,d\in\mathbb{C}$, let 
\[U(2)=\left(\begin{array}{cc}a & b \\c & d\end{array}\right)\]
such that $|a|^2+|c|^2=|b|^2+|d|^2=1$
\end{definition}

\subsection{Quantum Gates}

We have seen that classical computer circuits consist of $wires$ and $gates$: wires carries information along the circuits and logic gates perform manipulation of the information. The only non-trivial classical single bit logic gate is the $NOT$ gate, whose operation is defines by $0\mapsto 1$ and $1\mapsto 0$. We have introduced both a non-invertible  and an invertible version of such a gate; the matrix 
\[X=\left(\begin{array}{cc}0 & 1 \\1 & 0\end{array}\right)\]
implements the reversible not. Given a quantum bit
\[\alpha|0\rangle+\beta|1\rangle\]
we can represent the action of $X$ on it by 
\[X\left(\begin{array}{c}\alpha \\\beta\end{array}\right)=\left(\begin{array}{c}\beta \\\alpha\end{array}\right)\]
from which we see that $X$ extends naturally from classical to quantum bits. Unlike in the classical case, we have infinitely many more quantum gates for single qubits, since $S_2\subset U(2)$. Among the most important one-qubit gates we mention the {\sl Hadamard gate}
\[H=\frac{1}{\sqrt{2}}\left(\begin{array}{cc}1 & 1 \\1 & -1\end{array}\right)\]
and the gate 
\[Z=\left(\begin{array}{cc}1 & 0 \\0 & -1\end{array}\right)\]
\subsection{Multiple qubits gates}
A system with more than one qubit can be represented, again, by tensor products of single qubits.\\
Suppose, for instance we have two qubits 
\[|\psi_1\rangle=\alpha|0\rangle+\beta|1\rangle\]
and 
\[|\psi_2\rangle=\gamma|0\rangle+\delta|1\rangle\]
let us define 
\[S_2(\mathbb{C})=\{(x_0,x_1x_2, x_{3})\in\mathbb{C}^4|\ \sum_{i=0}^{3}|x_i|^2=1\}\]

Then we have:
\begin{theorem}
The tensor product $|\psi_1\rangle\otimes|\psi_2\rangle=^{def}|\psi_1\rangle|\psi_2\rangle$ belongs to $S_2(\mathbb{C}).$

\end{theorem}

$Proof.$ Since
\[|\alpha|^2+|\beta|^2=1\]
and
\[|\gamma|^2+|\delta|^2=1\]
we have that 
\[|\alpha\gamma|^2+|\alpha\delta|^2+|\beta\gamma|^2+|\beta\delta|^2=1\]
so that
\[=\alpha\gamma|00\rangle+\alpha\delta|01\rangle+\beta\gamma|10\rangle+\beta\delta|11\rangle\]
The above observation justifies the follow:
\begin{definition} The state space of a system of $n$ qubits is defined  by the algebraic curve lying in $\mathbb{C}^N$:
\[S_{N-1}(\mathbb{C})=\{(x_0,x_1,\dots x_{N-1})\in\mathbb{C}^N|\ \sum_{i=0}^{N-1}|x_i|^2=1\}\]
where $N=2^n$.  

\end{definition}
The linear group preserving such a set is the Unitary Group, so we can conclude that the geometry associated to Quantum Computing is represented by the pair $(S_{N-1}(\mathbb{C}),\mathbb{U}(N))$.\\
\begin{example}
The archetypal multi-qubit quantum logic gate is the $controlled-NOT$ or $CNOT$ gate. This gate has two input qubits, known as the control  and the target qubit. The matrix representation is the following
\[U_{CN}=\left(\begin{array}{cccc}1 & 0 & 0 & 0 \\0 & 1 & 0 & 0 \\0 & 0 & 0 & 1 \\0 & 0 & 1 & 0\end{array}\right)\]
and its action can be described as follows: if the the control qubit is set to 0, then nothing happens to the target; if the control qubit is set to 1, then the target qubit is flipped:
\[|00\rangle\mapsto|00\rangle;|01\rangle\mapsto|01\rangle;|10\rangle\mapsto|11\rangle;|11\rangle\mapsto|10\rangle\]
\end{example}

$U_{CN}$ is clearly unitary and captures the {\sl IF-THEN} rule. 

Finally, we remind the remarkable fact that, like for CQ, also for QC  it is possible to show that there exists a small library of $quantum\ gates$ that can simulate an arbitrary quantum computation:

\begin{theorem}
There exists a universal sets of gates for quantum computation.
\end{theorem}
 $Proof.$\  See $\cite{nc}.$
 \begin{flushright}
 $\square$
 \end{flushright}

Since we have established that any classical computation can be simulated by a permutation (which, in turn, is represented by a permutation matrix, which is unitary), then we can conclude that any classical computation can be simulated by a quantum computation.\\
\subsection{More on QC}
The biggest difference between classical and quantum computing lies in the very different way we can gain access to information. We can $measure$ the state of a classical system by simply looking at it, and getting as output the state in which it lives at the moment of the measurement.

Unlike for classical systems, by a postulate of quantum mechanic, we are not allowed to $observe$ $quantum\ states$, that is, we do not have direct access to the information encoded in some generic point of $S_{N-1}(\mathbb{C})$. Indee, if we perform a $measurement$, then the state we are observing $collapses$ into some classical state, that is an element of $\mathcal{B}(N)\subset S_{N-1}(\mathbb{C})$.

Ultimately, the only information we can actually read is still boolean; we remark the fact that the collapses is a projection operator, therefore  in particular is not linear.
The classical state which the quantum state collapses to is determined probabilistically by the distribution of probability written in the quantum state itself; more precisely, if we try to measure the qubit
\[|\psi\rangle=\alpha|0\rangle+\beta|0\rangle\]
we will end up with the state $|0\rangle$ with probability $|\alpha|^2$ and with the state $|1\rangle$ with probability $|\beta|^2$. 

This fact reflects the  non deterministic nature of quantum computing: we never have the absolute certainty of getting a certain classical state  out of quantum state unless such a quantum state is itself a classical one. 
\subsection{Complexity}

The {\sl complexity} (both in the CQ and QC) of a circuit can be defined as the number of gates of the circuit itself. One of the breakthroughs in QC has been the discover \cite{sh} that some problems can be solved exponentially faster on a Quantum than on a Classical computer. 
Deutsch \cite{de} has been one the pioneers in this fiels we asked wether it is possible for a quantum computer to efficiently solve computational problems which have no efficient solutions on a classical computer. He presented a simple example suggesting that, indeed, quantum computers might have computational powers exceeding those of classical computers.

After ten years Peter Shor \cite{sh} proved that two important problems- the problem of factoring integers and the so called `discrete logarithm' problem- could be solved efficiently on a quantum computer. This was a big achievement because these two problem do not seem to have an efficient solution of classical computers. In 1995 there was another sign showing the, at least theoretical, power of QC, when Lov Grover \cite{gr} showed that the problem of conducting a search through an unstructured database would have also speeded up by a quantum computer. Shor's celebrated algorithm relies essentially on the Quantum Fourier Transform (QFT) for cyclic groups which runs exponentially faster on quantum computers.

More recently David Rockmore \cite{ro} has extended the (QFT) to abelian groups, and his result has been used to prove some generalizations of Simon's problem \cite{si}, a class of problem named Hidden Subgroup Problems; which whereas the general question of the existence of a QFT for non abelian groups is still open.

One of the reason why we think that can be useful to introduce, as we shall do soon, the concept of {\sl fuzzy bit}, is that to have a common language in order to be able to compare the efficiency of solving certain class of problems with respect to QC and FC. 
 So, for instance, we can ask if the existence of a Fuzzy Fourier Transform (FFFT- not to confuse with FFT, often used to denote the Fast Fourier Transform) is a meaningful question to work with, and if so, if it can be framed in the model we are going to propose.
 
\section{Structures}
It can be noticed that the three geometries just described exhibit a very strong duality between the underlying linear object $\mathcal{O}$ where the information is encoded and the linear group acting naturally on $\mathcal{O}$. In the classical case, for instance, we have seen that $\mathcal{B}(N)$ is the canonical basis of $\mathbb{R}^N$, therefore its elements have the form:
\[(0,0,\dots,0,1,0,\dots,0)\]

The elements of correspondent group of transformation look much like its geometric counterpart; indeed an element of $P(N)$ has the property that, for any pair $(i,j)$, there is only one non-zero entry on the $i$-th row and the $j$-th column. These matrices have the form:
\[\left(\begin{array}{cccccc}0 & 1 & \dots & 0 & 0 & 0 \\0 & 0 & \dots & 0 & 0 & 0 \\\dots &  &  & 1 & 0 & 0 \\1 & 0 & \dots & 0 & 0 & 0 \\\dots & 0 & \dots & 0 & 1 & 0 \\0 & 0 & \dots & 1 & 0 & 1\end{array}\right)\]

The same phenomena happen for the pair $(Pr(N),St(N))$: in that case the points on the curve have the property that their coordinate add up to $1$, whereas the elements of $St(N)$ are matrices whose columns add up to 1.

As to the pair $(S_{N-1}(\mathbb{C}),U(N))$, the point on $S_{N-1}(\mathbb{C})$ have the property that the sum of the square of the modules of the coordinates add up to one, whereas an unitary matrix has the very analogous property that the sum of the square of the modules of the entries on each column (and each row) add up to one. In this case we also have the property that the columns (and the rows) of an unitary matrix form an orthonormal set.

These phenonema are an indication that there is a close relation between a linear object and the linear structures of matrices acting on it. The reason why we stress this point is that it has been one the leading motivation behind the ideas exposed in the next section.

\section{The model}

In the light of the above discussion, it would be desirable if fuzzy computing were a linear model too. More specifically, is there any consistent algebraic way to model the concept like `fuzzy bit' or `fuzzy gates'? That would give us, among other things, an effective method to compare the power of fuzzy computing with respect to other models of computation.
 
Of course, one of the main issues in fuzzy computing is the nature of the object where the truth values are stored: the unit interval is not an algebraic variety and it carries an algebraic structure very different from its classical or quantum counterpart. 

The role of weak structures is emerging as one of the newest and most promising `revolution' in modern mathematics and a great deal of work on semirings and its applications (see, for an introduction, \cite{go1} and \cite{go2}) to several fields of mathematics, logic and computer science has been done in recent years. 

This suggets that, nonetheless, a sort of linear calculus is possible also on [0,1], and, more in general, on MV-algebras, since to any an MV-algebra $A$ can be given a semiring structure, and we can use the  theory of semimodules to define a semiring of matrices acting on it as a semiring over a semimodule. 

In what follows we define a {\sl fuzzy bit} trying to follow as close as possible the definition given for quantum and classical bit. Then we define a new semiring of matrices which can be thought as the set of  {\sl fuzzy gates}, we study its structure and we use tensor products of semimodules to define a state space which naturally represents a system of multiple fuzzy bits. 

 We start our investigation by defining the objects where fuzzy information could be stored.

\subsection{Basic definitions}

\begin{definition}
A semiring $(R,+,\cdot)$ is a set $R$ with two binary operation $+,\cdot$ such that
\begin{enumerate}
\item $(R,+)$ is a commutative monoid with identity element 0;
\item $(R, \cdot)$ is a monoid with identity element 1$\neq 0$;
\item $a(b + c)=ab + ac$ and $(a + b)c = ac + bc$ for all $ a, b, c\in R$;
\item $0a =0=a0$ for all $a\in R$
\end{enumerate}
\end{definition}

\begin{example}

\begin{enumerate}

\item $(\mathbb{N}^+,+,\cdot)$, $(\mathbb{Q}^+,+,\cdot)$, $(\mathbb{R}^+,+,\cdot)$ are semiring, where + and $\cdot$ are the ordinary addition and multiplication, and as usual , $\mathbb{N},\mathbb{Q},\mathbb{R}$ are the non-negative natural, rational and real numbers, respectively. 
\item The {\sl fuzz semiring} {\sl ($\mathcal{I}$,max,min)}, where $\mathcal{I}=[0,1]$
\item The {\sl Viterbi semiring} $(\mathcal{I},max,\cdot)$
\item Given any semiring $(R,+,.)$ we can give a semiring structure to the set of the functions $R^X$ from any set $X$ to $R$ by defining addition and multiplication on $R^X$ pointwise:
\[ [f+g](x)=f(x)+g(x)\ \ \ [f\cdot g](x)=f(x)\cdot g(x)\]
\end{enumerate}
\end{example}

An element $r$ of a semiring $R$ is said to be {\sl idempotent} if $r+r=r$; the set of idempotent elements of a semiring $R$ is denoted by $I^+(R)$; if $I^+(R)=R$ then $R$ is said to be an $idempotent\ semiring$. In current literature, such structures are also called $diods$ os $semilattice\ ordered\ monoids$.

A semiring is called {\sl zerosumfree} if and only if $a+b=0\leftrightarrow a=b=0$; it is called {\sl entire} if and only if $ab=1\leftrightarrow a=b=1$. A semiring which is both entire and zerosumfree is called an {\sl information\ algebra}.

Additively idempotent semirings have been studied in several contests, and they are becoming an essential tool in modern mathematics; according to Litvinov and Maslov \cite{masl}, 

`If we have an important and interesting numerical algorithm, then there is a good chance that its semiring analogs are important and interesting as well'.

We say that a semiring $R$ is {\sl partially ordered} if and only if there is a partial order $\leq$ on $R$ satisfying the following conditions:
\begin{enumerate}
\item $a\leq b\Rightarrow a+c\leq b+c$
\item $a\leq b\ and\ \ 0\leq c\Rightarrow ac\leq bc\ and\ ca\leq cb$
\end{enumerate}
for all $a,b,c\in R$
 
 A semiring $R$ is positive if and only if it is partially-ordered and $0\leq a$ for all $a\in R$.\\
It is easy to check that any idempotent semiring $R$ is partially ordered by 
\[a\leq b\Leftrightarrow a+b=b\]
and that, under such an equivalence, $R$ is positive.\\

 Given a MV algebra $(A,\oplus,0_A,1_A,\neg)$ a MV define a `sum'
\[x\wedge y=x\odot(\neg x\oplus y)\]
It can be shown that $\wedge$ is distributive with respect to $\oplus$ 
\[x\oplus (y\wedge z)=(x\oplus y)\wedge (x\oplus z)\]

It is easy to check that the structure $(A,\wedge,\oplus,1,0)$ is a semiring: the sum is idempotent, because $x\vee x =(x\odot \neg x)\oplus x=x$, we have an additive identity is $1$, since $x\wedge 1=x\  \forall x\in A$, a multiplicative identity is $0$, since $0\oplus a=a\ \forall a\in  A$; we  also have a multiplicative zero, since $a\oplus 1=1\ \forall a\in A$.

In particular, the  MV-algebra $\mI=([0,1],\wedge,\oplus,1,0)$ is a semiring with `multiplication'
\[x\oplus y=min(x+y,1)\]
and `sum'
\[x\wedge y=min(x,y)\]

 $(\mI,\wedge,\oplus,1,0)$ is both zerosumfree and entire, therefore it is an example of an information algebra.
 
 The next notion we need to introduce is that of {\sl semimodule}:
 
\begin{definition}
A {\sl left-semimodule} $M$ over a semiring $(R,+,\cdot)$   is an abelian monoid $(M,+)$ with identity $0_M$  and an action $R\times M\rightarrow M$  denoted by $(r,m)\rightarrow rm$ called {\sl scalar  multiplication}, such that , for any $r,r'\in R$ and $m,m'\in M$
\begin{enumerate}
\item $(rr')m=r(r'm)$
\item $r(m+m')=rm+rm'$
\item $(r+r')m=rm+r'm$
\item $1 m=m$
\item $r 0_M=0_M=0m$
\end{enumerate}
\end{definition}

A right-semimodule is defined similarly; unless otherwise specified,  we will assume $R$ to be commutative, so that $M$ is both a right and left $R$-semimodule, so we can refer to it simply as a $R-$ semimodule.

Let us notice that, in particular, a commutative semiring $R$ is a semimodule on itself, with the semiring multiplication playing the role of the scalar multiplication.

In Di Nola \cite{dn} the concepts of linear independence, basis and homomorphisms are studied in the realm of semimodules, and it is shown that such concepts and many of the properties they have in a linear contests still hold for semilinear spaces; in what follows we give a list of some of them.

\begin{definition}
Let $M$ be a semimodule over a semiring $R$. A finite set $\{m_1,m_2,\dots,m_r\}$ of elements of M is said to be linearly independent if
\[\sum_{i=1}^nr_im_i=\sum_{i=1}^nr'_im_i\Rightarrow r_i=r'_i \]
where $r_i,r'_i\in R.$
\end{definition}

\begin{definition}
A basis for M is a linearly independent set of generators $\{m_1,m_2,\dots,m_r\}$ for M.
\end{definition}
Since we have basis, the concept of $rank$ is well defined for free semimodules: 
\begin{definition}
M is said to be a free of rank $r$ if it admits a basis $\{m_1,m_2,\dots,m_r\}.$
\end{definition}

\begin{example}
We have seen that  ($\I,\wedge,\oplus,1,0$), which we relabel as $(\I,+,\cdot,1,0)$ is a semiring. Define
\[\I^n=\{(a_1,a_2,\dots,a_n)\ |\ a_i\in\I,\forall i\}\]
Then we can make $\I^n$ into a $\I-$semimodule by componentwise addition and multiplication by elements of $\I.$ The module $\I^n$ is called $the\ free\ semimodule\ of\ rank\ n\ over\ \I.$ 
The canonical basis $\mathcal{B}$ for the free semimodule $\I^n$ is $\{e_1,e_2,\dots,e_n\}$ where $e_i$ is a $n-dimensional$ vectors with all the entries equal to $0$ except the $i-th$ one.
\end{example}
In particular, any semiring $R$ is free when considered as a semimodule on itself.
It is also possible to show that free semimodules are an universal object, so that they can be defined equivalently as follows:
\begin{definition}
For any set $A$ there is a free-semimodule $F(A)$ over the semiring $R$ on the set $A$ with the following $universal\ property$: if $M$ is any $R-$semimodule and $\phi:A\longrightarrow M$ is any map of sets, then there is a unique $R-semimodule$ homomorphism $\Phi:F(A)\longrightarrow M$ such that $\Phi(a)=\phi(a)$, for all $a\in A$, that is the following diagram commutes:
\begin{equation}
\xymatrix{ A\ar[r]^{i} \ar[dr]_{\phi} & F(A)\ar[d]^{\Phi}\\\
                                                      & M  }
\end{equation}
\end{definition}
The universal property of free semimodules is derives as corollary from one of the papers written by Katsov on the tensor product of semimodules \cite{yk2}.\\
If  $A$ is a finite set $\{a_1,a_2,\dots,a_n\}$, $F(A)=Ra_1\bigoplus Ra_2\bigoplus\dots,\bigoplus Ra_n\cong R^n$.

\subsection{Towards a semilinear fuzzy computing}

We propose a model where the elementary `fuzzy' bit of information is encoded in the semilinear object defined below. 
Let $\mathcal{I}=([0,1],\wedge,\oplus,1,0)$.

\begin{definition}
Let $S(2)=\{(a,b)\in \mathcal{I}\times \mathcal{I}\ | \ a\wedge b=0\}\cup\{(1,1)\}$.
\end{definition}
Let $\neg(a,b)=(1-a,1-b)$, and
\[(x,y)\circ(x',y')=(x\wedge x',y\wedge y').\]
 Then 

\begin{theorem}
$(S(2),\circ)$ is a commutative semigroup with involution and identity  $(1,1)$ 
\end{theorem}

$Proof$. Let $(x,y),(x',y')\in S(2)$. Then $(x,y)\circ(x',y')=(x\wedge x',y\wedge y')$. But then $(x,y)\circ(x',y')\in S(2)$ since $(x\wedge x')\wedge (x\wedge y')=x\wedge x'\wedge y\wedge y'=0$ (by hypothesis). $\wedge$ is associative by definition and, for any $(x,y)\in S(2)$ we have that $(x,y)\wedge (1,1)=(x\wedge 1,y\wedge 1)=(x,y)$.
\begin{flushright}
$\square$
\end{flushright}
\begin{definition}
Let us call {\sl fuzzy bit} an element of $S(2)$, and {\sl fuzzy set} a function $f:X\longrightarrow S(2)$, where $X$ is any set.
\end{definition}

Given a semiring $R$, the family of all the $n\times n$ matrices $M_n(R)$ is still a semiring, under componentwise addition and the ordinary matrix multiplication. The additive identity is the matrix with all its entries equal to 0 and the identity is the diagonal matrix with only 1 on the diagonal. If $R$  is idempotent also $M_n(R)$ has the same property and the partial order on $R$ induces a componentwise partial order  on $M_n(R)$. Motivated by this facts, we are going to define a subring of $M_2(\mathcal{I})$ which acts naturally, and linearly, on the set defined above. 
\begin{definition}
Let $MV_2(\I)$ be the set of $2\times 2$ matrices 
\[\left(\begin{array}{cc}a & b \\c & d\end{array}\right)\cup\left(\begin{array}{cc}1 & 1 \\1 & 1\end{array}\right)\]

with $a,b,c,d\in\I$ such that $a\wedge c=0$ and $b\wedge d=0$
\end{definition}
\begin{remark}
This mimics the one given for the stochastic matrices $a+b=1$
considering the fact that the `addition' in the MV semiring $S$ is the lattice operation $\wedge$ and the `multiplication` is the `truncated sum' $\oplus$, so that the unit with respect to the multiplication is $0$ in the same way 1 is the unit with respect to the multiplication between real numbers. 
\end{remark}

Notice that there are $(2^2-1)^2=9$ `types' of columns with the required property, so that we have $9$ types of such matrices, namely 
{\small \[\left\{\left(\begin{array}{cc}0 & 0 \\0 & 0\end{array}\right),\left(\begin{array}{cc}0 & 0 \\0 & a\end{array}\right),\left(\begin{array}{cc}0 & 0 \\a & 0\end{array}\right),\left(\begin{array}{cc}0 & 0 \\a & b\end{array}\right),\left(\begin{array}{cc}0 & a \\0 & 0\end{array}\right),\left(\begin{array}{cc}a & 0 \\0 & 0\end{array}\right),\left(\begin{array}{cc}a & b \\0 & 0\end{array}\right),\left(\begin{array}{cc}0 & a \\b & 0\end{array}\right),\left(\begin{array}{cc}a & 0 \\0 & b\end{array}\right)\right\}\]}
with $a,b\in \I$.\\

Let us define:
\begin{definition}
\begin{enumerate}
\item component-wise sum $\wedge$: $(A\wedge B)_{ij}=A_{ij}\wedge B_{ij}$ 
\item row by column product $\circ$: let  $A=\left(\begin{array}{cc}a & b \\c & d\end{array}\right)$ and  $B=\left(\begin{array}{cc}x & w \\ y & z\end{array}\right)$ then
\[A\circ B=\left(\begin{array}{cc}(a\oplus x)\wedge(b\oplus y) & (a\oplus w)\wedge(b\oplus z) \\(c\oplus x)\wedge(d\oplus y) & (c\oplus w)\wedge(d\oplus z)\end{array}\right)\]
\end{enumerate}
\end{definition}

Then we have the following:

\begin{lemma}
For any $A,B\in MV_2(\I)$, 
\begin{enumerate}
\item $A\circ B \in MV_2(\I)$
\item $A\circ(B\wedge C)=A\circ B\wedge A\circ C$ for any $A,B,C\in MV_2(\I).$
\end{enumerate}
\end{lemma}
$Proof.$ Let $A$ and $B$ as above. By the expression for $A\circ B$ we get the following for the sum of the elements in the first column: 
\[a\oplus x\wedge b\oplus y\wedge c\oplus x\wedge d\oplus y=\]
\[=(a\wedge c)\oplus x\wedge(b\wedge d)\oplus y=(0\oplus x)\wedge (0\oplus y)=0\oplus(x\wedge y)=0\oplus 0=0\] 
so that $A\circ B\in MV_2(\I)$.

The same computation applies to the elements of the second column.

It is check that
\[\left(\begin{array}{cc}1 & 1 \\1 & 1\end{array}\right)\circ A=\left(\begin{array}{cc}1 & 1 \\1 & 1\end{array}\right)\]

 Let $C=\left(\begin{array}{cc}e & f \\g & h\end{array}\right)$; then we have 
\[A\circ(B\wedge C)=\left(\begin{array}{cc}a & b \\c & d\end{array}\right)\left(\begin{array}{cc}x\wedge e & y\wedge f \\w\wedge g & z\wedge h\end{array}\right)=\]
\[=\left(\begin{array}{cc}a\oplus[x\wedge e]\wedge b\oplus[w\wedge g] & a\oplus[y\wedge f]\wedge b\oplus[z\wedge h] \\ c\oplus[x\wedge e]\wedge d\oplus[w\wedge g] & c\oplus[y\wedge f]\wedge d\oplus[z\wedge h]\end{array}\right)=\]

\[=\left(\begin{array}{cc}a\oplus x\wedge a\oplus e \wedge b\oplus w\wedge b\oplus g & a\oplus y\wedge a\oplus f \wedge b\oplus z\wedge b\oplus h \\ c\oplus x\wedge c\oplus e\wedge d\oplus w\wedge d\oplus g & c\oplus y\wedge c\oplus f \wedge d\oplus z\wedge d\oplus h\end{array}\right)=\]
$A\circ B\wedge A\circ C.$
\begin{flushright}
$\square$
\end{flushright}

\begin{lemma}
There is an identity element 1 with respect to $\circ$ in $MV_2(R)$, an identity element 0 with respect to $\wedge$ such that $0\neq 1$ and 0 is also such that $0\circ A=0$ for any $A\in MV_2(R)$.

\end{lemma}
$Proof.$ Let us notice that 
\[1=\left(\begin{array}{cc}0 & 1 \\1 & 0\end{array}\right)\]
has the property that for all $(x,y)\in S(2) $ and 
\[A=\left(\begin{array}{cc}x & y \\w & z\end{array}\right)\]
\[\left(\begin{array}{cc}0 & 1 \\1 & 0\end{array}\right)\left(\begin{array}{c}x \\w\end{array}\right)=\left(\begin{array}{c}0\oplus x\wedge 1\oplus w \\1\oplus x \wedge 0\oplus w\end{array}\right)=\left(\begin{array}{c}x\wedge 1 \\1\wedge w\end{array}\right)=\left(\begin{array}{c}x \\w\end{array}\right)\]
Since 1 acts as an identity on the column of the elements of $MV_2(S)$ by the action $\circ$, then 1 is the identity element with respecto to $\circ.$

We can also notice that 

\[0=\left(\begin{array}{cc}1 & 1 \\1 & 1\end{array}\right)\]

is such that 
\[\left(\begin{array}{cc}1 & 1 \\1 & 1\end{array}\right)\circ A=\left(\begin{array}{cc}1\oplus x\wedge 1\oplus w & 1\oplus y\wedge 1\oplus z \\1\oplus x\wedge 1\oplus w & 1\oplus y\wedge 1\oplus z\end{array}\right)=\left(\begin{array}{cc}1 & 1 \\1 & 1\end{array}\right) =0\]
This shows $0\neq 1$ is the absorbing element of the multiplication.
\begin{flushright}
$\square$
\end{flushright}

\begin{lemma}

There exists an element $J\in MV_2(\I)$ that acts as an involution.
\end{lemma}
$Proof.$ If we define $J$ as
\[J=\left(\begin{array}{cc}1 & 0 \\0 & 1\end{array}\right)\]
we can see that 
\[ J\circ J=\left(\begin{array}{cc}1 & 0 \\0 & 1\end{array}\right)\circ\left(\begin{array}{cc}1 & 0 \\0 & 1\end{array}\right)=\left(\begin{array}{cc}1\wedge 0 & 1\wedge 1 \\1\wedge 1 & 0\wedge 1\end{array}\right)=\left(\begin{array}{cc}0 & 1 \\1 & 0\end{array}\right)=I\]
\begin{flushright}
$\square$
\end{flushright}

The above lemmas show that:
 \begin{theorem}
$(MV_2(\I),\wedge,\circ)$ is a semiring with an element $J$ acting as an involution.
\end{theorem}
\begin{flushright}
$\square$
\end{flushright}

We are now interested to study the action of $MV_2(S)$ on $S(2)$.
Let us consider the natural row by column' action  of $MV_2(\I)$ on $S(2)$:
\[\left(\begin{array}{cc}a & b \\c & d\end{array}\right)\left(\begin{array}{c}x \\y\end{array}\right)=\left(\begin{array}{c}(a\oplus x)\wedge(b\oplus y) \\(c\oplus x)\wedge(d\oplus y)\end{array}\right)\]
We have:
\begin{lemma}
$A(v)\in S(2)$ for any $A=\left(\begin{array}{cc}a & b \\c & d\end{array}\right)\in MV_2(\I)$ and any $v=(x,y) \in S(2).$
\end{lemma}

$Proof.$\ Using the distributivity in $\I$:
\[[(a\oplus x)\wedge(b\oplus y)]\wedge[(c\oplus x)\wedge(d\oplus y)]=
[(a\wedge c)\oplus x]\wedge[(b\wedge d)\oplus y]=0 \]
We also have that 
\[\left(\begin{array}{cc}1 & 1 \\1 & 1\end{array}\right)\circ\left(\begin{array}{c}x \\y\end{array}\right)=(1,1)\in S(2)\]

\begin{flushright}
$\square$
\end{flushright}
The action just defined is linear on $S(2)$, that is, for any $A\in MV_2(\I)$ and $v,v'\in S(2)$, we have 
\begin{lemma}
$A(v\wedge v')=Av\wedge Av'$
\end{lemma}

$Proof.$ Let $A$ and $v$ as above and $v'=(x,x')$. We have:
\[A(v\wedge v')=\left(\begin{array}{cc}a & b \\c & d\end{array}\right)\left[\left(\begin{array}{c}x \\y\end{array}\right)\wedge\left(\begin{array}{c}x' \\y'\end{array}\right)\right]=
\left(\begin{array}{cc}a & b \\c & d\end{array}\right)\left(\begin{array}{c}x\wedge x' \\y\wedge y'\end{array}\right)=\]
\[\left(\begin{array}{c}(a\oplus (x\wedge x'))\wedge(b\oplus (y\wedge y')) \\(c\oplus (x\wedge x'))\wedge(d\oplus (y\wedge y'))\end{array}\right)=\]

\[=\left(\begin{array}{cc}(a\oplus x)\wedge (a\oplus x') & (b\oplus y)\wedge (b\oplus y') \\(c\oplus x)\wedge (c\oplus c') & (d\oplus x')\wedge(d\oplus y')\end{array}\right)=\]\[=A\left(\begin{array}{c}x \\y\end{array}\right)\wedge A \left(\begin{array}{c}x'\\y'\end{array}\right)=Av\wedge Av'\]
\begin{flushright}
$\square$
\end{flushright}

Let us now prove what is probably the most desirable property in the perspective of building up our linear machinery: we want to show that 
the action of $MV_2(\I)$ on $S(2)$ is the action of a semiring on a semimodule. This amounts to prove that:
\begin{lemma} 
\[(A\circ B)(v)=A[B(v)]\]
for any $A,B\in MV_2(\I)$ and any $v\in S(2)$. \end{lemma} 
$Proof$. Let $A,B$ as before and $v=(\alpha,\beta)$; we have that
\[A\circ[B(v)]=\left(\begin{array}{cc}a & b \\c & d\end{array}\right)[\left(\begin{array}{cc}x & y \\x & z\end{array}\right)\left(\begin{array}{c}\alpha \\\beta\end{array}\right)]=\]
\[=\left(\begin{array}{cc}a & b \\c & d\end{array}\right)\left(\begin{array}{c}x\oplus\alpha\wedge y\oplus \beta \\w\oplus \alpha\wedge z\oplus \beta\end{array}\right)=\]
\[=\left(\begin{array}{c}a\oplus (x\oplus \alpha\wedge y\oplus \beta)\wedge b\oplus (w\oplus \alpha\wedge z\oplus \beta) \\c\oplus (x\oplus \alpha\wedge y\oplus \beta)\wedge d\oplus (w\oplus \alpha\wedge z\oplus \beta)\end{array}\right)=\]
\[=\left(\begin{array}{c}(\alpha(a\oplus x\wedge b\oplus w))\wedge (\beta(a\oplus y\wedge b\oplus z)) \\(\alpha(c\oplus x\wedge d\oplus w))\wedge (\beta(c\oplus y\wedge d\oplus z))\end{array}\right)=\]
\[=(A\circ B)(v)\]
\begin{flushright}
$\square$
\end{flushright}

After noticing that, for any $(x,y)\in S(2)$
\[\left(\begin{array}{cc}1 & 1 \\1 & 1\end{array}\right)\circ\left(\begin{array}{c}x \\y\end{array}\right)=\left(\begin{array}{c}1 \\1\end{array}\right)\]
and
\[\left(\begin{array}{cc}0 & 1 \\1 & 0\end{array}\right)\circ\left(\begin{array}{c}x \\y\end{array}\right)=\left(\begin{array}{c}x \\y\end{array}\right)\]
we can gather together the above results to conclude that:

\begin{theorem} 
$S(2)$  is a semimodule over the semiring $MV_2(\I)$
\end{theorem}
\begin{flushright}
$\square$
\end{flushright}

\subsection{Tensor product of idempotent semimodules}

At first glance our choice of representing a fuzzy bit could seem obscure. A truth value in fuzzy logic is just an element of $\mathcal{I}$ or, more abstractly, an element of a MV-algebra; why did we decide to add a dimension and represent a fuzzy bit as a two dimensional object? 

The explanation is found in the algebraic properties of the tensor product, which we will discuss shortly; the idea is that, as we have seen for the case of CQ and QC, in order to represent systems of multiple fuzzy bit of information, we need higher dimensional objects in which to encode information.

As we shall see, tensor products of free semimodules are free objects of dimension equal to the {\sl product} of the dimension of the factors, so we must represent information with at least a 2-dimensional object if we want to work with systems of multiple fuzzy bits. 

Our main sources of theoretical results about tensor product of semimodules over idempotent semirings are Katsov \cite{yk1},\cite{yk2}, and Litvinov \cite{masl}; their work has been important to us in the sense that it set a solid theoretic ground to the algebraic structures we are interested in. For sake of clarity, in what follows we briefly recall the arguments developed by Katsov to prove the existence of a tensor product in the category of the semimodules over semirings; we refer to the unsurpassed treatise of S. Mac Lane \cite{sm} for the notions of category theory we are going to mention. 
 \begin{definition}
 An element $a$ of a monoid $S$ is said to be regular if $a=asa$ for some $s\in S$; $S$ is $regular$ if all its elements are regular. 
  \end{definition}
  \begin{definition}
  Given $a,b\in S$ , we say that $b$ is an inverse of $a$if $a=aba$ and $b=bab$. A monoid in which every element has a unique inverse is said an $inverse\ monoid$.
  \end{definition}
If $S$ is commutative, then $S$ is regular if and only of $S$ it is inverse.

Let $\mathcal{M}$ the category of commutative inverse monoid., and let $R_\mathcal{M}$ and $\mathcal{M}_R$ the category of left and right semimodules over an additively regular semiring.\\ 
It is possible to show that $\mathcal{M}$ is abelian and, by a result proved in Katsov \cite{yk2}, does have an internal tensor product bifunctor $-\otimes-:\mathcal{M}\times\mathcal{M}\rightarrow\mathcal{M}$. Moreover, the category $R_\mathcal{M}$ and $\mathcal{M}_R$ can be regarded as the categories of covariant and controvariant additive funcotrs from $R$ (considered as a category with one single object) to $\mathcal{M}$. $\mathcal{M}$ can be proved to be cocomplete, and therefore is provided with coproducts.

 All these observations can be used to conclude that constructions and result about the internal tensor product bifunctor $-\otimes-:\mathcal{M}\times\mathcal{M}\rightarrow\mathcal{M}$ can be extended to semimodules over additively regular semirings, which includes idempotent semirings. 
\begin{definition}
Let $F$ and $G$ respectively a right and a left semimodule on an additively regular semiring $R$. Then the tensor product $F\otimes_R G$ is defined as the factor monoid of the free semigroup $\mathcal{F}(F\times G)/\sim$, where $\sim$ is the congruence generated by the pairs $(xr\otimes y),(x\otimes ry)$, $x\in F $,$y\in G, r\in R$
\end{definition}

The previous construction is rather abstract; in \cite{masl} the explicit construction for idempotent semimodules over idempotent semirings is given, and a brief outline of which is given below for sake of clarity. In what follows we make the further assumption that semimodules are $algebraically\ complete$, the definition of which is given below. 
For the proof of next few theorems we refer to \cite{masl},\cite{yk1},\cite{yk2}. 

Assuming $S$ is a commutative semiring, algebraic completeness is defined as follows:

\begin{definition}
\begin{enumerate}
\item an idempotent semigroup is said to be {\sl algebraically complete} (a-complete) if each subset $X\subset S$ has the least upper bound.
\item an idempotent semiring $(S,\oplus,\otimes)$ is said to be {\sl a-complete} if it is $(S,\oplus)$ is an a-complete semigroup and, for any subset $X\subset S$ and any $s\in S$, we have  $s\otimes(\oplus X)=\oplus(k\otimes X)$
\item an idempotent semimodule $M$ over $S$ is said to be $a-complete$ if it contains the element $sup\ M$and  $(\oplus X)\otimes m=\oplus (Q\otimes x)$ for any $Q\in S$ and $m\in M$
\end{enumerate}
\end{definition}
\begin{definition}
 Given idempotent semimodules $M,N$\ and $P$ over an idempotent ring $(R,\oplus,\odot)$, we say that $f:M\rightarrow N$ is linear if $f(x\oplus y)=f(x)\oplus f(y)$ and $f[r\odot (x)]=r\odot f(x)$ for all $r\in R$. A map $b:M\times N\rightarrow P$ is said to be $bilinear$ if it is linear in each of the two component separately.
 \end{definition}

Consider the direct product $M=M_1\times M_2$ of two $a-complete$ idempotent semimodules $M_1,M_2$, both over a ring $S$ ; let $X\subset M$, $\vec{x}=(x_1,x_2)\in X$ and $a_{\vec{x}}$ an arbitrary function from $X$ to $S$ and let $T$ be the set of formal sums of the following kind:
\[t=\bigoplus_{\vec{x}\in X}a_{x_1,x_2}\odot (x_1\otimes x_2) \]

where $\odot$ is the multiplication in the semiring $S$.\\
Notice that $\oplus$ is the formal sum we have assigned to $T$ and does not have anything to do with the semiring addition; in particular, we can notice that such a formal sum can be represented by the union, therefore it is an idemp otent binary operation. In this way $T$ can be given the structure of semimodule on $S$, with scalar multiplication $\odot$. 

Let us introduce the following equivalence relation $\sim$ in $T$: 
\[k\odot(x_1\otimes x_2)=((k\odot x_1)\otimes x_2)=(x_1\otimes (k\odot x_2))\]
or
\[(x_1\oplus x_2)\otimes x_3= (x_1\otimes x_3)\oplus(x_2\otimes x_3)\]
We denote $T\sim$ by $M_1\otimes M_2$;  $M_1\otimes M_2$ is a `good' tensor product, in the sense that the following universal property holds:

\begin{theorem} 
The canonical projection 
\[\beta_{M_1,M_2}:M_1\times M_2\longrightarrow M_1\otimes M_2\]
\[(x_1,x_2)\mapsto x_1\otimes x_2\]
is bilinear. Moreover, given an arbitrary  idempotent semimodules $W$,  for any bilinear map $b:M_1\times M_2\longrightarrow W$ there is a unique bilinear map 
\[\tilde{b}:M_1\otimes M_2\longrightarrow W\] 
such that 
\[f=\tilde{b}\circ \beta_{M_1,M_2}\]
\end{theorem}
\begin{flushright}
$\square$
\end{flushright}

We call $M_1\otimes M_2$ the tensor product (over $R$) of $M_1$ with $M_2$.

Let $M,N,P$ be semimodules over a commutative idempotent semiring $R$. Let us denote by $M\bigoplus N$ the direct sum of $M$ and $N$; then we have that the following properties of the tensor product just defined hold. The following properties of the tensor product of idempotent semimodules hold:

\begin{theorem}

\begin{enumerate}
\item $M\otimes N =N\otimes M$
\item $(M\bigoplus N)\otimes P=M\otimes P\bigoplus N\otimes P$
\item $(M\otimes N)\otimes P=M\otimes(N\otimes P)$
\end{enumerate}
\end{theorem}

\begin{flushright}
$\square$
\end{flushright}

Let us $(R,\oplus,\odot)$ a commutative idempotent semiring, $X$ an arbitrary set and $\mathcal{B}(X,R)$ the set of bounded functions on $X$. $B(X,R)$ is a  $R$-semimodule under the pointwise semiring operations. In particular if we chose $R$ to be $\mathcal{I}$ then we have the following 
\begin{theorem}
$\mathcal{B}(X,\mathcal{I})\otimes\mathcal{B}(Y,\mathcal{I})$ and 
$\mathcal{B}(X\times Y,\mathcal{I})$ are isomorphic.
\end{theorem}

\begin{flushright}
$\square$
\end{flushright}

As before, let $\mathcal{M}$ be the category of commutative inverse semimodules, $\mathcal{M}R$ and $_R\mathcal{M}$ the category of right and left semimodules over the semiring $R$, $R_R$ and $_RR$ semiring $R$ seen as right and left semimodule on itself, $Id$
 the forgetful tensor from $_R\mathcal{M}$ and $\mathcal{M}_R$ to $_R$, then we have 
  \begin{theorem} 

For functors $-\otimes  {_RR} : \mathcal{M}_R\longrightarrow \mathcal{M}$, and $ R_R\otimes- :_R\mathcal{M}\longrightarrow \mathcal{M}$, there are natural isomorphisms: $-\otimes{_R R} \cong Id$, and $R_R\otimes- \cong Id$. 
\end{theorem}
\begin{flushright}
$\square$ 
\end{flushright}
Now we can show that:
\begin{theorem}
Let $R$ a commutative semiring and let $M\cong R^s$ and $N\cong R^t$ be free $R-modules$ with bases $m_1,m_2,\dots,m_s$ and $n_1,n_2,\dots,n_t$ respectively. Then $M\otimes_R N$ is a free $R-module$ of rank $st$ with basis $m_i\otimes n_j$, $1\leq i\leq s$ and 
$1\leq j\leq t$, i.e.
\[R^s\otimes R^t\cong R^{st}\]

\end{theorem}
$Proof:$\ Let $M=m_1R\oplus m_2R\oplus\dots\oplus m_sR$ and $N=n_1R\oplus n_2R\oplus\dots\oplus n_tR$. Then, by Thm. 24 
\[M\otimes N=m_1R\otimes n_1R\oplus m_1R\otimes n_2 R\oplus \dots\oplus m_1R\otimes n_tR\oplus m_2R\otimes n_1 R\oplus \dots\oplus m_sR\otimes n_tR\]

From Thm. 26, 
\[m_iR\otimes n_jR=(m_i\otimes n_j)R\]

 by which the theorem follows.
\begin{flushright}
$\square$
\end{flushright}

This result makes clear that tensor product of free idempotent semimodules behaves well with respect to the dimensions of the component; motivated by this fact we are now able to introduce the tensor product of homomorphisms of semimodules: let $M,M',N,N'$ be two semimodules over a semiring $R$ and suppose 
\[\phi:M\longrightarrow M'\] and 
\[\psi:N\longrightarrow N'\] 
are $R$-semimodule homomorphisms. Then the `tensor product of homomorphisms' is well defined:
\begin{theorem}
There is a unique $R-semimodule$ homomorphism, denoted by $\phi\otimes \psi$, mapping $M\otimes N$ into $M'\otimes N'$ such that $(\phi\otimes\psi)(m\otimes n)=\phi(m)\otimes\psi(n)$ for all $m\in M$ and $n\in N$. 
\end{theorem}

Since $MV(n)$ acts linearly on $S(n)$, then a matrix $M\in MV(n)$ represents a $MV(n)-$ semimodule homomorphism of $S(n)$ in itself with respect to some chosen basis, which we assume to be the canonical basis  $\mathcal{B}$.

As before, let $V,W,X,Y$ be free semimodule of finite rank over the same semiring $R$ and let $\phi:V\longrightarrow X$ and $\psi:W\longrightarrow Y$ be two $R-$ homomorohisms (of semimodules).
\begin{theorem}
We can represent
\[\phi\otimes\psi:V\otimes W\longrightarrow X\otimes Y\]
by the Kronecker product of two matrices representing $\psi$ and $\phi$.
\end{theorem}

$Proof.$\ Let $\mathcal{B}_1=\{v_1,v_2,\dots,v_n\}$ and $\mathcal{B}_2=\{w_1,w_2,\dots,w_m\}$ be ordered bases of $V$ and $W$ respectively, and let $\mathcal{E}_1=\{x_1,x_2,\dots,x_r\}$  and $\mathcal{E}_2=\{y_1,y_2,\dots,y_s\}$ be ordered basis of $X$  and $Y$ respectively.  Then $\mathcal{B}=v_i\otimes w_j$ and $\mathcal{E}=x_i\otimes y_j$ are the basis of $V\otimes W$ and $X\otimes Y$. Suppose 
\[\phi(v_i)=\sum_{p=1}^r\alpha_{pi}x_p\ \ \ and\ \  \ \psi(w_j)=\sum_{q=1}^s\beta_{qi}y_q\]
Then
\begin{equation}
(\phi\otimes\psi)(v_i\otimes w_j)=(\phi(v_i))\otimes(\psi(w_j))=\]
\[=(\sum_{p=1}^r\alpha_{pi}x_p)\otimes(\sum_{q-1}^s\beta_{qi}y_q)=\sum_{p=1}^r\sum_{q=1}^s\alpha_{pi}\beta_{qj}(x_p\otimes y_q)
\end{equation}

Since we can order the basis $\mathcal{E}$ and $\mathcal{B}$ respectively in $r$ and $s$ ordered sets, the above equation specifies the entries for the correspondent matrix of $\phi\otimes\psi$. The resulting matrix $M_{\mathcal{B}}^{\mathcal{E}}(\phi\otimes\psi)$ is an $r\times n$ block matrix whose $p,q$ block is the $s\times m$ matrix $\alpha_{p,q}M_{\mathcal{B}_2}^{\mathcal{E}_2}(\psi).$  
\begin{flushright}
$\square$
\end{flushright}

\subsection{Systems of multiple fuzzy bits}
  The canonical basis for $\mI\times\mI$ is $\mathcal{B}_2=\{|0\rangle=(0,1),|1\rangle=(1,0)\}$. Indeed we can see that $|0\rangle$ and $|1\rangle$ are linearly independent and that, given any $(x,y)\in S(2)$, $(x,y)=x(0,1)\wedge y(1,0)=x|0\rangle\wedge y|1\rangle$, where, in order to simplify the notation, we have replaced $a\oplus b$ by $ab$. Since $S(2)$ is a subsemimodule  of $\mI\times\mI$, let us look now at $S(2)\otimes S(2)\subset \mI^2\otimes\mI^2\cong\mI^4$.\\
Let us start with two fuzzy bits: $\psi_1=(a,b)$ and $\psi_2=(c,d)$; since a Kronecker product between vectors has been defined in the previous section, let us compute $\psi_1\otimes\psi_2$

  \[\psi_1\otimes \psi_2=(ac,ad,bc,bd)\]
Therefore 
\[|00\rangle=|0\rangle\otimes|0\rangle=(0,1)\otimes(0,1)=(0,1,1,1)\]
\[|01\rangle=|0\rangle\otimes|1\rangle=(0,1)\otimes(1,0)=(1,0,1,1)\]
\[|10\rangle=|1\rangle\otimes|0\rangle=(1,0)\otimes(0,1)=(1,1,0,1)\]
\[|11\rangle=|1\rangle\otimes|1\rangle=(1,0)\otimes(1,0)=(1,1,1,0)\]
Let us extend linearly such product to all of $S(2)$; that is, given $\psi_1=(a,b)=a|0\rangle+b|1\rangle$ and $\psi_2=(c,d)=c|0\rangle|+d|1\rangle$ in $S(2)$, we can write  
\[\psi_1\otimes \psi_2= ac|00\rangle+ad|01\rangle+bc|10\rangle+bd|11\rangle=(ac,ad,bc,bd)\]
We notice that,  
\[ac\wedge ad\wedge bc\wedge bd=a(c\wedge d)\wedge b(c\wedge d)=0\] 

by hypothesis, therefore, if we define 

\[S(4)=\{(a_0,a_1,a_2,a_3)\in \mathcal{I}^4\ |\ \bigwedge_{i=0}^{3}a_i=0\}\]
we see that we have a map

\[S(2)\longrightarrow S(4)\]
\[(\psi_1,\psi_2)\mapsto (\psi_1\otimes \psi_2)\]
Let us now compute the Kronecker product between elements of $MV_2(S)$. Let 
\[A=\left(\begin{array}{cc}a & b \\c & d\end{array}\right)\]
 and 
 \[B=\left(\begin{array}{cc}x & y \\w & z\end{array}\right)\]
Then 
\[A\otimes B=\left(\begin{array}{cc}aB & bB \\cB & dB\end{array}\right)\]
By the same argument as above, it follows that the `sum'  of the entries of the columns of $A\otimes B$ is 0: 
\[\bigwedge_{j=1}^4 a_{i,j}=0,\ \ \ \forall i=1,2,3,4\]
Therefore, if we define

\[MV_4(S)=\left\{ \left(\begin{array}{cccc}a_{11} & \dots &  & a_{14} \\a_{22} & \dots &  &  \\\dots &  &  &  \\a_{41} & \dots &  & a_{44}\end{array}\right)  \ |\ \bigwedge_{j=1}^4 a_{i,j}=0,\ \ \ \forall i=1,2,3,4\right\}\]

we have a map 
\[MV_2(S)\times MV_2(S)\longrightarrow MV_4(S)\]
\[(A,B)\mapsto A\otimes B\]
Now we can generalize our construction by defining 
\[S(N)=\{(a_0,a_1,\dots,a_{N-1})\ |\ \bigwedge_{i=0}^{N-1}a_i=0\}\]

\[MV_{N}(S)=\left\{ \left(\begin{array}{cccc}a_{11} & \dots &  & a_{1N} \\a_{22} & \dots &  &  \\\dots &  &  &  \\a_{N-1} & \dots &  & a_{NN}\end{array}\right)  \ |\ \bigwedge_{j=1}^N a_{i,j}=0,\ \ \ \forall i=1,2,\dots,N\right\}\]

The operators we have defined appear to be good candidates to play the role of {\sl fuzzy gates} in fuzzy computing and we have plans to test the robustness and the efficience of our model in the next future.\\ We would like to conclude with a few questions related to the ideas exposed in this work:
\begin{itemize}
\item what kind of computational problems can be coded and eventually solved using this model?
\item is it possible technically implement and concretely to encode, and treat,  information in the way we have just described?
\item are there algorithms that can be computed on a machine based on such a model more efficiently than using boolean or quantum computing?

\end{itemize} 

The particular choice of the semirings $MV_n(S)$ as main actor of our model is clearly arbitrary, even thought has been motivated by some analogies with stochastic computing. A negative answer to some, or even all of the questions above, does not rule out {\sl a priori} the possibility of finding semirings  by which an efficient fuzzy computing can be build: the semiring itself, as well as the semimodule which defines fuzzy bits, can be considered as a sort of free variable, describing an entire family of variations of our model. In this sense we would like to think that our contribution is directed more towards the big picture (framing fuzzy computing as a particular case of a more general theory of computation) rather than associated with the choice of a particular algebraic object.


\begin{thebibliography}{99}
   
\bibitem{Zah} Zadeh, L. A., Fuzzy sets, \emph{Information and Control}, {\bf 8} (1965), 338-353.
\bibitem{fe1} Feynman, R., Simulating physics with computers, \emph {Int. J. Theor. Phys.}, {\bf 21} (1982), 467-488.
\bibitem{fe2} Feynman, R., \emph{Feynman Lectures on Computation}, Addison-Wesley, Boston, MA, 1996.
\bibitem{tu} Turing, A., On computable numbers, with an application to the Entscheidungsproblem, \emph {Proc.  Lond. Math. Soc.}, (2) {\bf 42} (1936), 230-265.
\bibitem{ft} Fredkin, E. and Toffoli, T., Conservative logic, \emph{Ont. J. Theor. Physics}, (3/4) {\bf 21} (1982), 219-253.
\bibitem{nc} Nielsen, M. A.  and Chaung, I. L.,  \emph{Quantum Computation and Quantum Information}, Cambridge University Press, Cambridge, UK, 2000.
\bibitem{sh}  Shor, P., Algorithms for quantum computation: discrete logarithms and factoring, \emph{Proceedings, Symposium on Foundations of Computer Science}, Santa F\'e, New Mexico, 1994.
\bibitem{de} Deutsch, D., Uncertainty in quantum measurements, \emph{Pys. Rev. Lett.}, (9) {\bf 50} (1983), 631-633.
\bibitem{gr} Grover, L., Quantum mechanic helps in searching for a needle in a haystack, \emph{Phys. Rev. Lett.}, (2) {\bf 79} (1997), 325.
\bibitem{ro} Rockmore, D., Fast Fourier analysis for abelian group extensions,  \emph{Adv. in Appl. Math.}, {\bf 11} (1990), 164-204. 
\bibitem{si} Simon, B., On the power of quantum computing, \emph{Proceedings, Symposium on Foundations of Computer Science}, Santa F\'e, New Mexico, 1997.
\bibitem{go1} Golan, J. S., \emph{Semirings and their applications},  Springer, New York-Heidelberg-Berlin, 1999.
\bibitem{go2} Golan, J. S., \emph{Power Algebras over Semirings},  Springer, New York-Heidelberg-Berlin, 1999.
\bibitem{masl} Litvinov, G. L., Maslov, V. P.  and Shpiz, G. B., Tensor product of idempotent semimodules, an algebraic approach, \emph{arXiv: math}, FA/0101153 v1,  2001.
\bibitem{dn} Di Nola, A.,  Lettieri, A., Nov\'ak, V. and  Perfilieva, I., Algebraic aspects of fuzzy systems,  \emph{Preprint}, (2005).
\bibitem{yk2}  Katsov, Y., Tensor product and injective envelopes of semimodules over additively-regular semirings. \emph{ Algebra Colloquium }, (2) {\bf 4} (1997), 121-131. 
\bibitem{yk1}  Katsov, Y., Toward homological characterization of semirings: Serre's conjecture and Bass's perfectness in a semiring context, \emph{Algebra univers.}, {\bf 52} (2004), 197-214.
\bibitem{fe1} Feynman, R., Simulating physics with computers, \emph {Int. J. Theor. Phys.}, {\bf 21} (1982), 467-488.
\bibitem{fe2} Feynman, R., \emph{Feynman Lectures on Computation}, Addison-Wesley, Boston, MA, 1996.

\end{thebibliography}
\end{document}